# Improved Res2Net model for Person re-identification

Zongjing Cao
Division of Computer Science and Engineering,
Jeonbuk National University
Jeonju City, Korea
caozongjing@gmail.com

Hyo Jong Lee*
Division of Computer Science and Engineering,
Jeonbuk National University
Jeonju City, Korea
hlee@chonbuk.ac.kr

*Abstract*—Person re-identification has become a very popular research topic in the computer vision community owing to its numerous applications and growing importance in visual surveillance. Person re-identification remains challenging due to occlusion, illumination and significant intra-class variations across different cameras. In this paper, we propose a multi-task network base on an improved Res2Net model that simultaneously computes the identification loss and verification loss of two pedestrian images. Given a pair of pedestrian images, the system predicts the identities of the two input images and whether they belong to the same identity. In order to obtain deeper feature information of pedestrians, we propose to use the latest Res2Net model for feature extraction of each input image. Experiments on several large-scale person re-identification benchmark datasets demonstrate the accuracy of our approach. For example, rank-1 accuracies are 83.18% (+1.38) and 93.14% (+0.84) for the DukeMTMC and Market-1501 datasets, respectively. The proposed method shows encouraging improvements compared with state-of-the-art methods.

## I. INTRODUCTION

Person re-identification (re-ID) is usually viewed as an image retrieval problem, which aims to match pedestrians from multiple cameras [1-3]. Given a person-of-interest (query), person re-ID determines whether the person has been observed by another camera [4]. Recent progress in this area has been flourished due to two factors: (1) the availability of the large-scale pedestrian datasets and (2) the learned pedestrian descriptor using a convolutional neural network (CNNs). Presently, deep learning methods dominate this community, with convincing superiority against hand-crafted competitors. Deeply-learned representations provide high discriminative ability, especially when aggregated from deeply-learned part features [5]. The latest state of the art on re-ID benchmarks are achieved with part-informed deep features.

Hand-crafted part features for person retrieval. Before deep learning methods dominated the re-ID research community, hand-crafted algorithms had developed approaches to learn part or local features. Gray and Tao [6] partition pedestrians into horizontal stripes to extract color and texture features. Similar partitions have then been adopted by many works [7, 8]. Some other works employ a more sophisticated strategy. Gheissari et al. [9] divided the pedestrian into several triangles for part feature extraction. Cheng et al. [10] employed a pictorial structure to parse the pedestrian into semantic parts. Das et al. [11] applied HSV histograms on the head, torso, and legs to capture spatial information.

Deeply-learned part features. The state of the art on most person retrieval datasets is presently maintained by deep learning methods [1]. When learning part features for re-ID, the advantages of deep learning over hand-crafted algorithms are two-fold. First, deep features generically obtain stronger discriminative ability. Second, deep learning offers better tools for parsing pedestrians, which further benefits the part features. In particular, human pose estimation and landmark detection have achieved impressive progress [12]. Several recent works in re-ID employ these tools for pedestrian partition and report encouraging improvement [13]. However, the underlying gap between datasets for pose estimation and person retrieval remains a problem when directly utilizing these pose estimation methods in an off-the-shelf manner. Others abandon the semantic cues for partition. Yao et al. [14] cluster the coordinates of max activations on feature maps to locate several regions of interest. Both Liu et al. [15] and Zhao et al. [16] embed the attention mechanism [17] in the network, allowing the model to decide where to focus by itself.

Recently, the CNNs has shown potential for learning state-of-the-art feature embeddings or deep metrics [4, 18-20]. Verification models and identification models are two major types of CNNs model structures in person re-ID. The two models are different concerning input, feature extraction, and loss function for training. Many previous works treat person re-ID as a binary classification task [3, 21-23] or a similarity regression task [23]. However, the major problem in the verification models is that they only use weak re-ID labels (same/different) and do not take all of the annotated information into consideration. Therefore, the verification network lacks the ability to consider the relationship between the image pairs in the datasets and other images. In the attempt to take full advantage of the re-ID labels, the identification models take a single image as input $x$ and predict the predefined identity label $t$. The main disadvantage of the identification model is that the training objective is different from the testing process, that is, it does not account for the similarity measurement between pairs of images, which may cause problems during the pedestrian retrieval process. In order to combine the advantages of the two models to improve the discriminative ability of the re-ID system, we propose a multi-task network based on Res2Net models.

The rest of this paper is organized as follows: In Section II, we describe the structure of our proposed network. In Section III, we describe the implementation details, proposed method, and experimental results. The conclusion is provided in section IV.

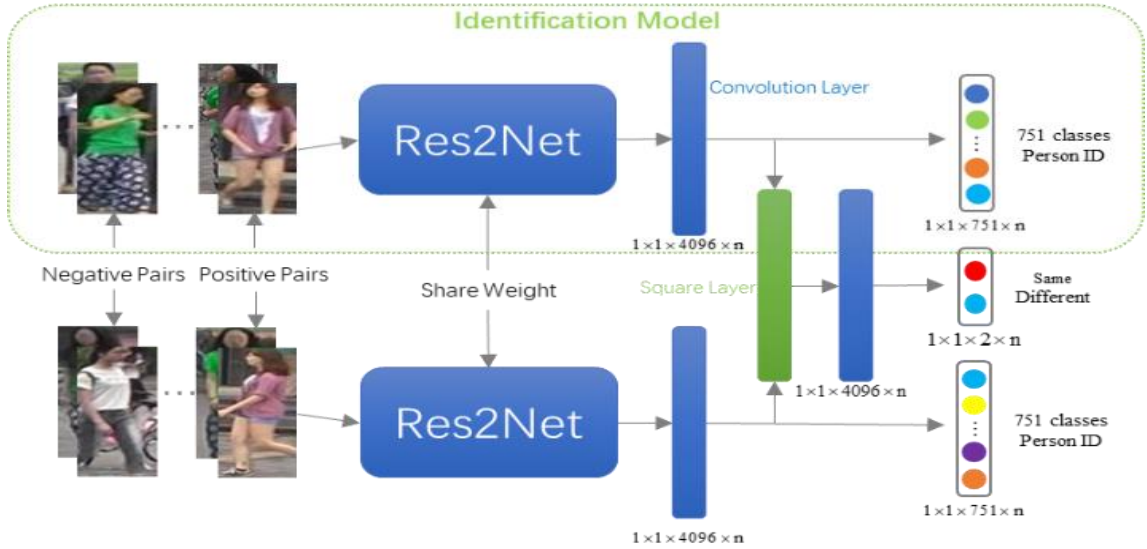

Fig. 1. A proposed Multi-Task Network model structure.

## II. PROPOSED METHOD

### A. Multi-task convolutional neural network

The proposed network is multi-task convolutional networks that combine two identification sub-networks and one verification sub-network. Fig. 1 briefly illustrates the architecture of the multi-task convolutional network. Given two images of resized to 256×128 as inputs, the multi-task network simultaneously predicts the identity label and the similarity score of the two input images. The multi-task convolutional network consists of two identification sub-networks based on ImageNet pre-trained Res2Net [24] models. The entire system is supervised by two identification loss and on verification loss. The implementation details of the model are described in section III.

### B. Res2Net: A New Multi-scale Backbone Architecture

Res2Net was a novel building block for CNNs, proposed by Shanghua Gao et al [24], by constructing hierarchical residual-like connections within one single residual block. The Res2Net represents multi-scale features at a granular level and increases the range of receptive fields for each network layer.

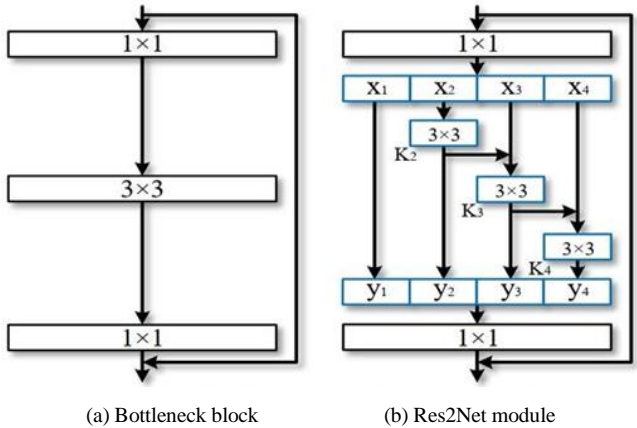

(a) Bottleneck block     (b) Res2Net module

Fig. 2. Comparison between the bottleneck block and the proposed Res2Net module (the scale dimension s = 4).

Different from some concurrent works [25-27] that improved the multi-scale ability by utilizing features with different resolutions; the multi-scale of our proposed method refers to the multiple available receptive fields at a more granular level. To achieve this goal, we replace the 3×3 filters of n channels, with a set of smaller filter groups, each with w channels (without loss of generality we use $n = s \times w$). As shown in Fig. 2, these smaller filter groups are connected in a hierarchical residual-like style to increase the number of scales that the output features can represent. Specifically, we divide input feature maps into several groups. A group of filters first extracts features from a group of input feature maps. Output features of the previous group are then sent to the next group of filters along with another group of input feature maps. This process repeats several times until all input feature maps are processed. Finally, feature maps from all groups are concatenated and sent to another group of $1 \times 1$ filters to fuse information altogether. Along with any possible path in which input features are transformed to output features, the equivalent receptive field increases whenever it passes a $3 \times 3$ filter, resulting in many equivalent feature scales due to combination effects.

### C. Identification Loss

The network has two Res2Net [24] models; they share weights and simultaneously predict two identity labels of the input images. Because the number of the training identities is 751 in Market1501 datasets, the convolutional layer has 751 kernels of size $1 \times 1 \times 4096$ connected to the output $f$ of Res2Net [24] and uses SoftMax unit to normalize the network output. The size of the output is $1 \times 1 \times 751$. We choose the cross-entropy loss for identity prediction, which is

$$\hat{p} = softmax(\theta_I \circ f), \qquad (1)$$

$$Identify(f, t, \theta_I) = \sum_{i=1}^{k} -p_i \log(\hat{p}_i). \qquad (2)$$

Where ∘ is a convolutional operator, $f$ is a $1 \times 1 \times 4096$ tensor, $t$ is the target class, and $\theta_I$ denotes the parameters of the convolutional layer. The $\hat{p}$ and $p_i$ are the predicted probability and the target probability, respectively, where $p_i = 0$ for all $i$ except $p_t = 1$.

*D. Verification Loss*

As shown in Fig. 1, we use a square layer to compare the features. The square layer takes $f_1$ and $f_2$ as input and $f_s$ is the output of the square layer. The square layer is denoted as

$$f_s = (f_1 - f_2)^2. \quad (3)$$

We treat pedestrian verification as a binary classification problem and use cross-entropy loss for predicted probability, which is

$$\hat{q} = soft\,max(\theta_s \circ f_s), \quad (4)$$

$$Verify(f_1, f_2, s, \theta_s) = \sum_{i=1}^{2} -q_i \log(\hat{q}_i). \quad (5)$$

Where the size of $f_1$ and $f_2$ is $1 \times 1 \times 4096$, $s$ is the target class (same/different), $\theta_s$ are parameters of the convolutional layer, and $\hat{q}$ is the predicted probability. If the inputs image depicts the same person, $q_1 = 1$, $q_2 = 0$; otherwise, $q_1 = 0$, $q_2 = 1$.

*E. Performance evaluation*

The mean average precision (mAP) and Cumulative Matching Characteristic (CMC) curve are two leading evaluation indicators used in the person re-ID task. mAP indicates the recall rate and is defined as follows:

$$mAP = \frac{\sum_{q=1}^{Q} AveP(q)}{Q}, \quad (6)$$

where Q is the number of queries in the set, and *AveP(q)* is the AP for a given query *q*.

CMC curve is an evaluation index to reflect retrieval accuracy and is defined as follows:

$$Acc_k = \begin{cases} 1 & \text{if } top-k \text{ ranked gallery samples contain the query identity} \\ 0 & \text{otherwise} \end{cases} \quad (7)$$

## III. EXPERIMENTS

*A. Datasets*

In this study, we used Market-1501 [23], DukeMTMC-reID [28, 29] datasets for evaluation of re-ID. The datasets are summarized in Table 1. Some images from these two datasets are shown in Fig. 3.

Market-1501: The Market-1501 is a benchmark dataset collected and annotated in Tsinghua campus, with 1501 identities, a total of 32668 annotated bounding boxes and each identity image is captured by at most six cameras [23]. According to the dataset setting, the training set contains 12,936 cropped images of 751 identities, and the testing set contains 19,732 cropped images of 750 identities and distractors [5]. For each query, we aim to retrieve the ground-truth images from the 19,732 candidate images.

DukeMTMC-reID: The DukeMTMC-reID dataset is collected on the Duke campus. The DukeMTMC-reID dataset is a subset of the DukeMTMC dataset for video-based person re-ID. The DukeMTMC-reID dataset contains 1,404 identities, 2,228 queries, 16,522 training images and 17,661 gallery images [28, 29]. With so many images captured by 8 cameras, DukeMTMC-reID manifests itself as one of the most challenging re-ID datasets up to now.

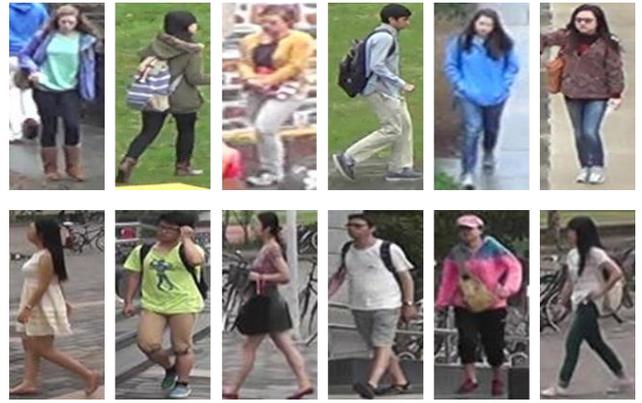

Figure 4. Sample images from DukeMTMC (Top) and Market-1501 (Bottom) datasets

Table 1. Summary of datasets for re-ID and attributes

| Dataset | Images | Gallery | Identities | Cameras |
|---|---|---|---|---|
| Market1501 | 12,936 | 19,732 | 1,501 | 6 |
| DukeMTMC | 16,522 | 17,661 | 1,404 | 8 |

*B. Implementation details*

Image preprocessing. The images have been randomly cropped to $256 \times 128$ and used in random order. Then we sample another image from the same/different class to compose a positive/negative pair.

Training. The number of training epochs is set to 256 for the proposed network. The initial learning rate was set as 0.05 and then decay learning rate by a factor of 0.1 every 40 epochs. For optimizer, we adopt the mini-batch Stochastic Gradient Descent to update the parameters of the multi-task network [30]. There are three types of loss in the proposed network. Therefore, we first compute all gradients produced by every loss respectively and add the weighted gradients together to update the network. We assign a weight 0.5 for the two gradients produced by two identification losses and weight 1 to the gradient produced by the verification loss.

Testing. Because two Res2Net share weights, the model has nearly the same memory consumption with the pre-trained model. Thus, we extract features by only activating one finetuned model. Given a $256 \times 128$ gallery image, we feed forward the image to one Res2Net [24] model in our multi-task network and get a pedestrian descriptor *f* of which size is $4096 \times 1$. Once the descriptors for the gallery are sets and then they are stored offline. Given a $256 \times 128$ query image, the network will extract its descriptor online and then sort the cosine distance between the query image and all gallery features to obtain the final ranking result.

*C. Some tricks for training model*

Random Erasing Augmentation. In person re-ID, persons in the images are sometimes occluded by other objects. To improve the generalization ability and address the occlusion problem of our models, we use the Random Erasing Augmentation (REA) [31] method to preprocessing the images before inputting the network [11]. Some examples are shown in Fig. 4. In training, the random erasing method randomly selects a rectangle region in the image and erases its original pixels using random values. In practice, we set the drop rate of REA to 0.5. In practice, we found that when REA

[31] is set to 0.4, the model can get the best rank-1 ranking and when set to 0.5, the mAP had the best result. The results are shown in Table 2.

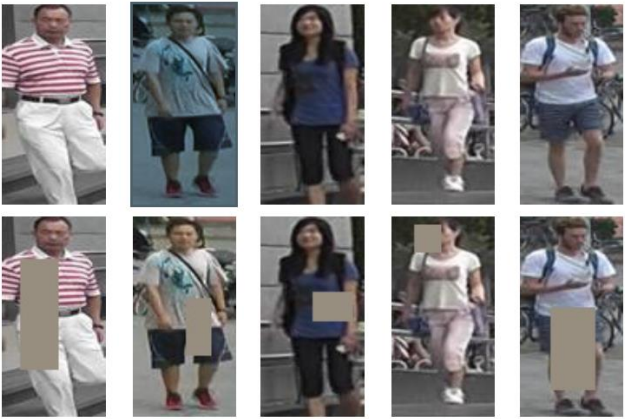

Fig. 4. Sampled examples of random erasing augmentation. The first row shows five original training images. The processed images are presented in the second low.

Table 2. Comparison with the different REA values on Market-1501 datasets

| REA | Rank-1 | mAP |
| --- | --- | --- |
| Er_0.0 | 88.24% | 71.14% |
| Er_0.1 | 89.82% | 75.10% |
| Er_0.2 | 90.05% | 75.43% |
| Er_0.3 | 89.96% | 75.50% |
| Er_0.4 | 90.47% | 76.13% |
| Er_0.5 | 90.02% | 76.39% |
| Er_0.6 | 89.46% | 74.75% |
| Er_0.7 | 89.04% | 74.58% |
| Er_0.8 | 89.01% | 74.06% |
| Er_0.9 | 89.16% | 74.31% |
| Er_1.0 | 89.19% | 74.14% |

**Warmup Learning Rate.** The learning rate has a great impact on the performance of a person re-ID model. Since the weights of the model are randomly initialized at the beginning of training, choosing a large learning rate at this time may bring instability to the model [30]. A warmup strategy is applied to bootstrap the network for better performance. In practice, we used a learning rate of 0.1 times for the first five epochs of the training network.

### D. Performance evaluation

The results on Market-1501 and DukeMTMC-reID datasets are shown in Table 4 which indicates the effectiveness of the proposed feature extracting strategy. In the experiment set, ten existing methods, include LOMO+XQDA [32], BoW+kissme [23], SVDNet [33], PAR [16], PAN [34], GAN [29], SVDNet+Era [31], MultiScate [35], PCB [36], MultiLoss [37] are tested on Market-1501 and DukeMTMC-reID datasets to compare with the proposed method. The proposed model using Res2Net-50 produces the best performance—92.93% in rank-1accuracy and 77.35% in mAP—which outperforms other state-of-the-art algorithms. As shown in Table 3, we also report the results obtained with the ResNet-50 and ResNet-121 networks, respectively, on Market-1501 and DukeMTMC-reID datasets. The experiment results on two different datasets have validated the proposed method better than other existing methods.

Table 3. Comparison of the proposed method with the ResNet-50 and ResNet-121 on Market-1501 and DukeMTMC-reID

| Methods | Market-1501 | | DukeMTMC-reID | |
| --- | --- | --- | --- | --- |
| | Rank-1 | mAP | Rank-1 | mAP |
| ResNet-50 | 90.47 | 76.13 | 80.71 | 65.31 |
| Res2Net-50 | **92.93** | **77.35** | **82.67** | **67.87** |
| ResNet-121 | 90.76 | 75.60 | 80.86 | 65.92 |
| Res2Net-121 | **93.14** | **77.67** | **83.18** | **68.16** |

Table 4. Comparison of the proposed method with the state-of-the-art methods on Market-1501 and DukeMTMC-reID

| Methods | Market-1501 | | DukeMTMC-reID | |
| --- | --- | --- | --- | --- |
| | Rank-1 | mAP | Rank-1 | mAP |
| LOMO+XQDA [16] | 43.79 | 22.22 | 30.8 | 17 |
| BoW+kissme [23] | 44.4 | 20.8 | 25.1 | 12.2 |
| SVDNet [33] | 82.3 | 62.1 | 76.7 | 56.8 |
| PAR [16] | 81.0 | 63.4 | - | - |
| PAN [34] | 82.8 | 63.4 | 71.6 | 51.5 |
| GAN [29] | 83.97 | 66.07 | 67.7 | 47.1 |
| SVDNet+Era [31] | 87.08 | 71.31 | 79.31 | 62.44 |
| MultiScate [35] | 88.9 | 73.1 | 79.2 | 60.6 |
| MultiLoss [37] | 83.9 | 64.4 | - | - |
| PCB [36] | 92.3 | 77.4 | 81.8 | 66.1 |
| Proposed | **93.14** | **77.67** | **83.18** | **68.16** |

**Instance Retrieval.** we apply the multi-task network to the generic pedestrian retrieval task. The results are shown in Fig. 5 and Fig. 6. The image in the leftmost is the query images. The retrieved images are sorted according to the similarity score from left to right. The label of the correctly matched image is drawn in green frames, and the label of the false matching image is in red frames.

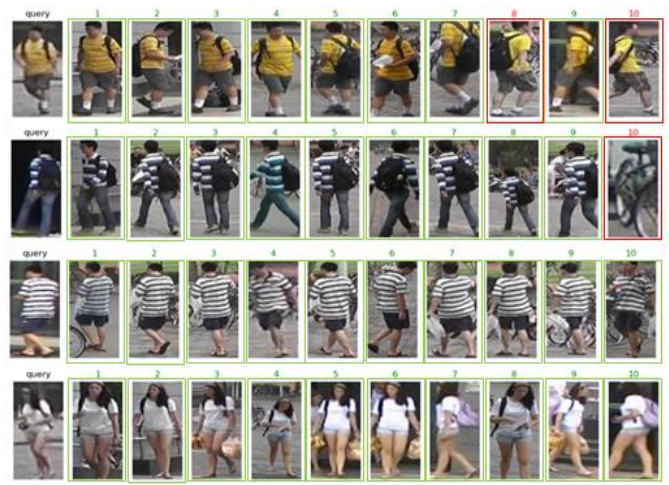

Fig. 5. Samples of pedestrian retrieval on the Market-1501 dataset

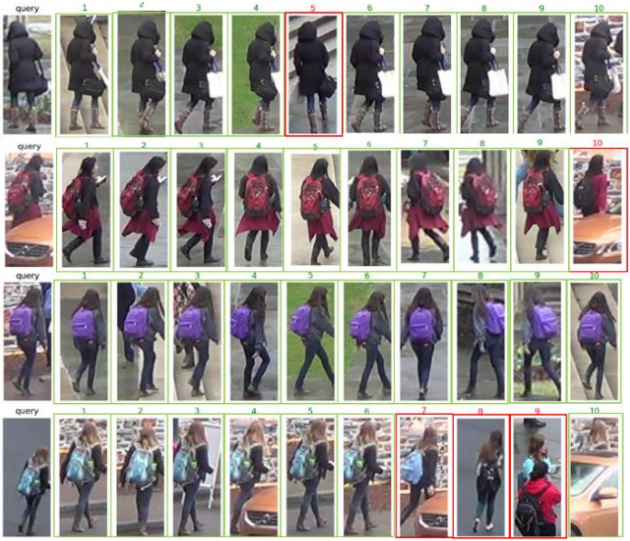

Fig. 6. Samples of pedestrian retrieval on the DukeMTMC-reID dataset

## IV. CONCLUSION

This paper contributed to solving the pedestrian re-ID problem. We proposed a multi-task network with a novel building block Res2Net that simultaneously considers verification loss and identification loss. The proposed model learns a discriminative embedding and a similarity measurement at the same time. The experimental results demonstrate the proposed method is better than other benchmarked methods. In the future, we will try to improve our network with better loss functions and data augmentation methods. Adapting this approach to video-based re-ID is another future avenue to invest.

ACKNOWLEDGMENT

This research was supported by Basic Science Research Program through the National Research Foundation of Korea (NRF) funded by the Ministry of Education (GR 2019R1D1A3A03103736).